\documentclass[]{article}
\usepackage{proceed2e}
\usepackage{graphicx}
\usepackage{amsmath, amsfonts}
\usepackage{tikz}

\usepackage[authoryear]{natbib}
\usepackage{hyperref}
\usepackage{amsthm}

\theoremstyle{definition}
\newtheorem{exmp}{Example}[section]

\usepackage{times}

\title{Bayesian Network Models for Adaptive Testing}

\author{ {\bf Martin~Plajner }\\
Institute of Information Theory and Automation\\
Academy of Sciences of the Czech Republic\\
Pod vod\'{a}renskou v\v{e}\v{z}\'{i} 4\\
Prague 8, CZ-182 08\\
Czech Republic\\
\And
{\bf Ji\v{r}\'{i}~Vomlel}  \\
Institute of Information Theory and Automation\\
Academy of Sciences of the Czech Republic\\
Pod vod\'{a}renskou v\v{e}\v{z}\'{i} 4\\
Prague 8, CZ-182 08\\
Czech Republic\\
}

\begin{document}

\maketitle

\begin{abstract}
Computerized adaptive testing (CAT) is an interesting and promising approach to testing human abilities. In our research we use Bayesian networks to create a model of tested humans.
We collected data from paper tests performed with grammar school students. In this article we first provide the summary of data used for our experiments. We propose several different Bayesian networks, which we tested and compared by cross-validation. Interesting results were obtained and are discussed in the paper. The analysis has brought a clearer view on the model selection problem.  Future research is outlined in the concluding part of the paper. 
\end{abstract}

\section{INTRODUCTION}
The testing of human knowledge is a very large field of human effort. We are in touch with different ability and skill checks almost daily. The computerized form of testing is also getting an increased attention with the growing spread of computers, smart phones and other devices which allow easy impact on the target groups. In this paper we focus on the Computerized Adaptive Testing (CAT)~\citep{CAT,Almond1999}. 

CAT aims at creating shorter tests and thus it takes less time without sacrificing its reliability. This type of test is computer administered. The test has an accompanied model which models a student (a student model). This model is constructed based on samples of previous students. During the testing the model is updated to reflect abilities of one particular student who is in the process of testing. At the same time we use the model to adaptively select next questions to be asked in order to ask the most appropriate one. This leads to collection of significant information in shorter time and allows to ask less questions. We provide an additional description of the testing process in the Section~\ref{sec-tests} and more information can be found also in~\citep{Millan2000}. It seems that there is a large possibility of applications of CAT in the domain of educational testing~\citep{VOMLEL2004a,Weiss1984}.

In this paper we look into the problem of using Bayesian network models~\citep{Kjrulff2008} for adaptive testing~\citep{Millan2010}. Bayesian network is a conditional independence structure and its usage for CAT can be understood as an expansion of the Item Response Theory (IRT)~\citep{Almond1999}. IRT has been successfully used in testing for many years already and experiments using Bayesian networks in CAT are also being made~\citep{Mislevy1994,VOMLEL2004}. 

We discuss the construction of Bayesian network models for data collected in paper tests organized at grammar schools. 
We propose and experimentally compare different Bayesian network models. To evaluate models we simulate tests using parts of collected data. Results of all proposed models are discussed and further research is outlined in the last section of this paper.
\section{DATA COLLECTION}
We designed a paper test of mathematical knowledge of grammar school students  focused on simple functions (mostly polynomial, trigonometric, and exponential/logarithmic). Students were asked to solve different mathematical problems\footnote{In this case we use the term mathematical ``problem'' due to its nature. In general tests,	 terms ``question'' or ``item'' are often used. In this article all of these terms are interchangeable.} including graph drawing and reading, calculation of points on the graph, root finding, description of function shape and other function properties. 

The test design went through two rounds. First, we prepared an initial version of the test. This version was carried out by a small group of students. We evaluated the first version of the test and based on this evaluation 
we made changes before the main test cycle. Problems were updated and changed to be better understood by students. Few problems were removed completely from the test, mainly because the information benefit of the problem was too low due to its high or low difficulty. Moreover we divided problems into subproblems in the way that: 
\begin{itemize}
\item[(a)] it is possible to separate the subproblem from the main problem and solve it independently or 
\item[(b)] it is not possible to separate the subproblem, but it represents a subroutine of the main problem solution. 
\end{itemize}
Note that each subproblem of the first type can be viewed as a completely separate problem. 
On the other hand, subproblems of the second type are inseparable pieces of a problem.

Next we present an example of a problem that appeared in the test.
\begin{exmp}
Decide which of the following functions 
\begin{eqnarray*}
f(x) & = & x^2-2x-8\\ 
g(x) & = & -x^2 +2x+8
\end{eqnarray*}
is decreasing in the interval $\left( -\infty,-1\right]$.
\end{exmp}

The final version of test contains 29 mathematical problems. Each one of them is graded with 0--4 points. 
These problems have been further divided into 53 subproblems.
Subproblems are graded so that the sum of their grades is the grade of the parent problem, i.e., it falls into the set $\{0,\ldots,4\}$. Usually a question is divided into two parts each graded by at most two points\footnote{There is one exception from this rule: The first problem is very simple and it is divided into 8 parts, each graded by zero or one point (summing to the total maximum of 8).}. The granularity of subproblems is not the same for all of them and is a subset of the set $\{0,\ldots,4\}$. All together, the maximal possible score to obtain in the test is 120 points. 
In an alternative evaluation approach, each subproblem is evaluated using the Boolean values (correct/wrong). The answer
is evaluated as correct only if the solution of the subproblem and the solution method is correct 
unless there is an obvious numerical mistake. 

We organized tests at four grammar schools. In total 281 students participated in the testing. In addition to problem solutions, we also collected basic personal data from students including age, gender, name, and their grades in mathematics, physics, and chemistry from previous three school terms. The primal goal of the tests was not the student
evaluation. The goal was to provide them valuable information about their weak and strong points. 
They could view their result (the scores obtained in each individual problem) as well as a 
comparison with the rest of the test group. The comparisons were provided in the form 
of quantiles in their class, school and all participants.

The Table~\ref{tab:score} shows the average scores of the grammar schools (the higher the score the better the results).
We also computed correlations between 
the score and average grades from Mathematics, Physics, and Chemistry from previous three school terms. 
The grades are from the set $\{1,2,3,4,5\}$ with the best grade being 1 and the worst being 5. These correlations are shown in the Table~\ref{tab:corel1}. Negative numbers mean that a better grade is correlated with a better result, which confirms our expectation. 


\begin{table}[htbp]
  \centering
  \caption{Average test scores of the four grammar schools.}
	
	\vspace{3mm}
	
    \begin{tabular}{|rrrr|r|}
		\hline
    GS1   & GS2 & GS3 & GS4 & Total \\
					\hline
    42.76 & 46.68 & 46.35 & 43.65 & 44.53 \\
		\hline
    \end{tabular}%
  \label{tab:score}%
\end{table}%

\begin{table}[htbp]
  \centering
  \caption{Correlation of the grades and the test total score.}

	\vspace{3mm}
	
    \begin{tabular}{|rrr|}
    \hline
    Mathematics & Physics & Chemistry \\
		\hline
    -0.60 & -0.42 & -0.41 \\
		\hline
    \end{tabular}%
  \label{tab:corel1}%
\end{table}%

%

\section{BAYESIAN NETWORK MODELS}\label{sec-models}

In this section we discuss different Bayesian network models we used to model relations between students' math skills 
and students' results when solving mathematical problems.
All models discussed in this paper consists of the following:
\begin{itemize} 
\item A set of $n$ variables we want to estimate $\{S_1,\ldots,S_n\}$. 
We will call them skills or skill variables.
We will use symbol $S$ to denote the multivariable $(S_1,\ldots,S_n)$ taking states $s = (s_1,\ldots s_n)$. 
\item A set of $m$ questions (math problems) $\{X_1,\ldots,X_m\}$.  
We will use the symbol $X$ to denote the multivariable $(X_1,\ldots,X_m)$ taking states $x = (x_1,\ldots,x_m)$.
\item A set of arcs between variables that define relations between skills and questions and, eventually, also 
inbetween skills and inbetween questions. 
\end{itemize}
The ultimate goal is to estimate the values of skills, i.e., the probabilities of states of variables $S_1,\ldots,S_n$. 

\subsection{QUESTIONS}

The solution of math problems were either evaluated using a numeric scale or using a Boolean scale as explained in the previous section. Although the numeric scale carries more information and thus it seems to be a better alternative, there are other aspects discouraging such a choice. The main problem is the model learning. The more the states the higher the number of model parameters to be learned. With a limited training data it may be difficult to reliably estimate the model parameters.

We consider two alternatives in our models. 
Variables corresponding to problems' solutions (questions) can either be
\begin{itemize}
\item Boolean, i.e. they have two states only $0$ and $1$ or 
\item integer, i.e. each $X_i$ takes $m_i$ states $\{1,\ldots,m_i\}$, $m_i \in \mathbb{N}$,
where $m_i$ is the maximal number points for the corresponding math problem.
\end{itemize} 
In Section~\ref{sec-evaluation} we present results of experiments with both options.

\subsection{SKILL NODES}

We assume the student responses can be explained by skill nodes that are parents of questions. 
Skill nodes model the student abilities and, generally, they are not directly observable.
Several decisions are to be made during the model creation.
 
The first decision is the number of skill nodes itself. Should we expect one common skill or should it rather be several different skills each related to a subset of questions only? In the later case it is necessary to specify
which skills are required to solve each particular question (i.e. a math problem). 
Skills required for the successful solution of a question become parents of the considered question.

Most networks proposed in this paper have only one skill node. This node is connected to all questions.
The student is thus modelled by a single variable. 
Ordinarily, it is not possible to give a precise interpretation to this variable. 

We created two models with more than one skill node. One of them is with the Boolean scale of question nodes and the other is with the numeric scale. We used our expert knowledge of the field 
of secondary school mathematics and our experiences gained during the evaluation of paper tests. 
In these model we included $7$ skill nodes with arcs connecting each of them to $1$ -- $4$ problems.

Another issue is the state space of the skill nodes. As an unobserved variable, it is hard to decide how many states it should have. Another alternative is to use a continuous skill variable instead of a discrete one but we did not elaborate more 
on this option. In our models we have used skill nodes with either $2$ or $3$ states
($s_i \in \{1,2\}$ or $s_i \in \{1,2,3\}$).

\label{observed_score}
We tried also the possibility of replacing the unobserved skill variable by a variable representing a total score of the test. To do this we had to use a coarse discretization. We divided the scores into three equally sized groups and thus we obtained an observed variable having three possible states. The states represent a group of students with ``bad'', ``average'', and ``good'' scores achieved. The state of this variable is known if all questions were included in the test. Thus, during the learning phase the variable is observed and the information is used for learning. On the other hand, during the testing the resulting score is not known -- we are trying to estimate the group into which would this test subject fall. In the testing phase the variable is hidden (unobserved).

\subsection{ADDITIONAL INFORMATION}

As mentioned above, we have collected not only solutions to problems but also additional personal information about students. This additional information may improve the quality of the student model. On the other hand it makes the model 
more complex (more parameters need to be estimated). It may mislead the reasoning based solely on question answers
(especially later when sufficient information about a student is collected from his/her answers). 
The additional variables are $Y_1,\ldots,Y_{\ell}$ and they take states $y_1,\ldots,y_{\ell}$.
We tested both versions of most of the models, i.e. models with or without the additional information.

\subsection{PROPOSED MODELS}

\begin{figure*}[htb]
\includegraphics[width=0.95\textwidth]{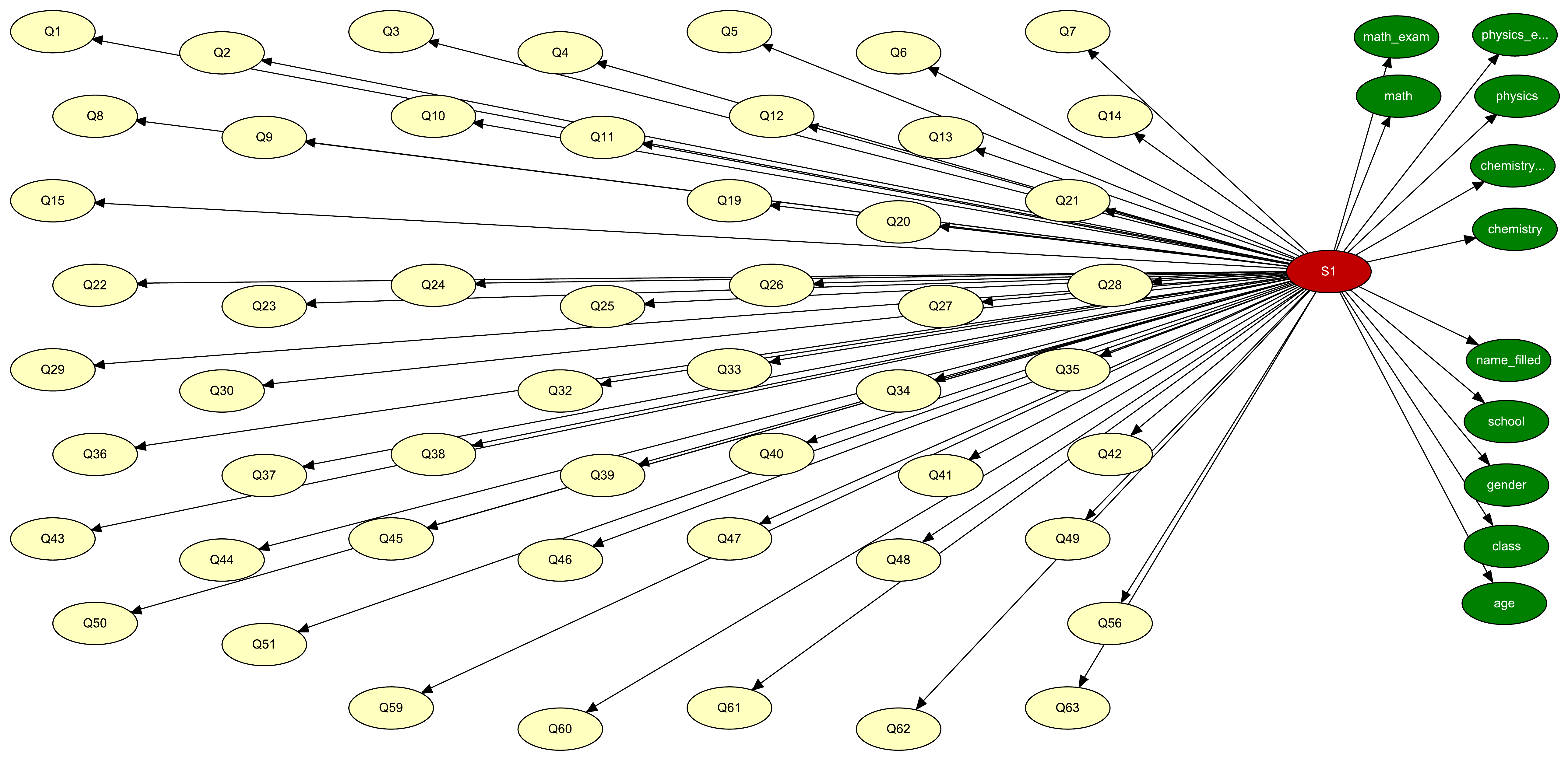}
\caption{Bayesian network with one hidden variable and personal information about students}
\label{net_1}
\end{figure*}

\begin{figure*}[htb]
\includegraphics[width=0.95\textwidth]{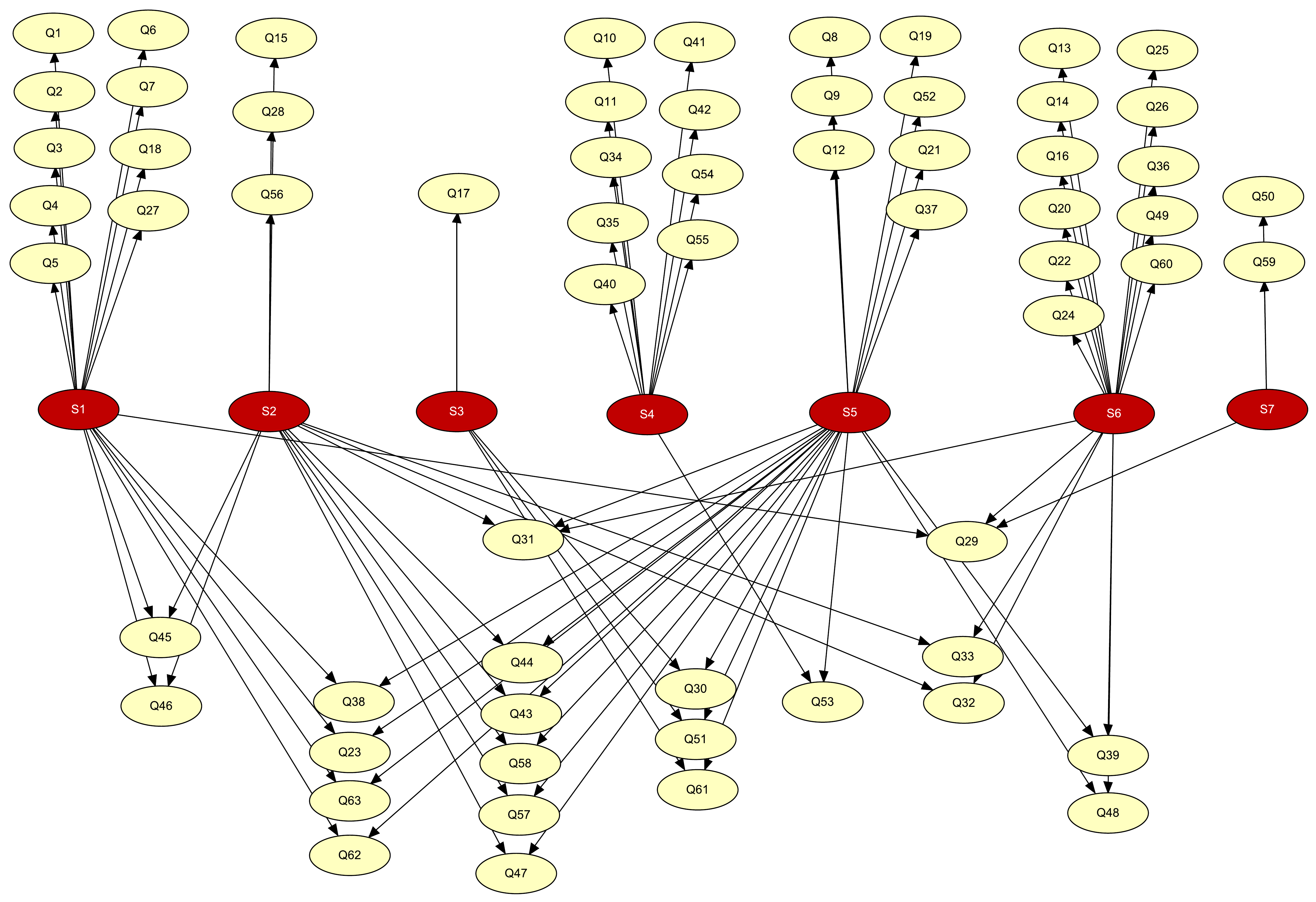}
\caption{Bayesian network with 7 hidden variables (the expert model)}
\label{net_2}
\end{figure*}

In total we have created $14$ different models that differ in factors discussed above. 
The combinations of parameters' settings are displayed in the Table~\ref{tab:BayesianNetworkModelsOverview}. 
One model type is shown in the Figure~\ref{net_1}. It is the case of "tf\_plus" which is a network with one hidden skill node and with the additional information\footnote{Please note that the missing problems and problem numbers are due to the two-cycled test creation and problems removal.}.
Models that differ only by number of states of variables have the same structure. Models with the ``obs'' infix in the name and ``o'' in the ID have the skill variable modified to represent score groups rather than skill (as explained earlier in the part~\ref{observed_score}).
Models without additional information do not contain the part of variables on the right hand side of the skill variable $S_1$. Figure~\ref{net_2} shows the structure of the expert models with $7$ skill variables in the middle part of the figure.

\begin{table}
	\centering
		\begin{tabular}{lccccc}
			ID & Model name & \rotatebox{90}{No. of skill nodes} & \rotatebox{90}{No. of states of skill nodes} & \rotatebox{90}{Problem variables} & \rotatebox{90}{Additional info} \\
			\hline
			b2 & tf\_simple & 1 & 2 & Boolean & no \\
			b2+ & tf\_plus & 1 & 2 & Boolean & yes \\
			b3 & tf3s\_simple & 1 & 3 & Boolean & no \\
			b3+ & tf3s\_plus & 1 & 3 & Boolean & yes \\
			b3o & tf3s\_obssimple & 1 & 3 & Boolean & no \\
			b3o+ & tf3s\_obsplus & 1 & 3 & Boolean & yes \\
			b2e &tf\_expert & 7 & 2 & Boolean & no \\
			
			\hline
			
			n2 & points\_simple & 1 & 2 & numeric & no \\
			n2+ & points\_plus & 1 & 2 & numeric & yes \\
			n3 & points3s\_simple & 1 & 3 & numeric & no \\
			n3+ & points3s\_plus & 1 & 3 & numeric & yes \\
			n3o & points3s\_obssimple & 1 & 3 & numeric & no \\
			n3o+ & points3s\_obsplus & 1 & 3 & numeric & yes \\
			n2e &points\_expert & 7 & 2 & numeric & no \\
			\hline
			\end{tabular}
	\caption{Overview of Bayesian network models}
	\label{tab:BayesianNetworkModelsOverview}
\end{table}

\section{ADAPTIVE TESTS}\label{sec-tests}

All proposed models are supposed to serve for adaptive testing. 
In this section we describe the process of adaptive testing with the help of these models. 

At first, we select the model which we want to use. If this model contains additional information variables it is necessary to insert observed states of these variables before we start selecting and asking questions. 
Next, following steps are repeated:
\begin{itemize}
	\item 	The next question to be asked is selected.
	\item 	The question is asked and a result is obtained.
	\item 	The result is inserted into the network as evidence.
	\item 	The network is updated with this evidence.
	\item 	(optional) Subsequent answers are estimated.
\end{itemize}

This procedure is repeated as long as necessary. It means until we reach a termination criterion which can be either a time restriction, the number of questions, or a confidence interval of the estimated variables. Each of these criterion would lead to a different learning strategy~\citep{VOMLEL2004}, but because such strategy would be NP-Hard~\citep{Lin2005}. We have chosen an heuristic approach based on greedy entropy minimization.

\subsection{SELECTING NEXT QUESTION}
One task to solve during the procedure is the selection of the next question. It is repeated in every step of the testing and it is described below.

Let the test be in the state after $s-1$ steps where 
\begin{eqnarray*}
\mathcal{X}_s & = & \{X_{i_1}\ldots X_{i_n} \ | \ i_1,\ldots,i_n \in \{1,\ldots,m\}\}
\end{eqnarray*}
are unobserved (unanswered) variables and 
\small
\begin{eqnarray*}
\lefteqn{e \ \ =}\\
 & & \{X_{k_1} = x_{k_1},\ldots,X_{k_o} = x_{k_o} | k_1,\ldots,k_o \in \{1,\ldots,m\}\} 
\end{eqnarray*}
\normalsize
is evidence of observed variables -- questions which were already answered and, possibly, the initial information. 
The goal is to select a variable from $\mathcal{X}_s$ to be asked as the next question. 
We select a question with the largest expected information gain. 

We compute the cumulative Shannon entropy over all skill variables of $S$ given evidence $e$.
It is given by the following formula:
\begin{eqnarray*}
H(e) & = & \sum_{i = 1}^n \sum_{s_i} -P(S_i=s_i|e) \cdot \log P(S_i=s_i|e) \enspace .
\end{eqnarray*}

Assume we decide to ask a question $X' \in \mathcal{X}_s$ with possible outcomes $x'_1,\ldots,x'_p$. 
After inserting the observed outcome the entropy over all skills changes. 
We can compute the value of new entropy for evidence extended by $X' = x_j'$, $j \in \{1,\ldots,p\}$ as:
\begin{eqnarray*}
\lefteqn{H(e,X'=x_j') \ \ =}\\
& & \sum_{i =1}^n \sum_{s_i} \begin{array}{ll}
-P(S_i=s_i|e,X'=x_j')\\
\cdot \log P(S_i=s_i|e,X'=x_j') 
\end{array}\enspace .
\end{eqnarray*}
This entropy $H(e,X'=x_j')$ is the sum of individual entropies over all skill nodes. Another option would be to compute the entropy of the joint probability distribution of all skill nodes. This would take into account correlations between these nodes. In our task we want to estimate marginal probabilities of all skill nodes. In the case of high correlations between two (or more) skills the second criterion would assign them a lower significance in the model. This is the behavior we wanted to avoid. The first criterion assigns the same significance to all skill nodes which seems to us as a better solution. Given the objective of the question selection, the greedy strategy based on the sum of entropies provides good results. Moreover, the computational time required for the proposed method is lower.

Now, we can compute the expected entropy after answering question $X'$: 
\begin{eqnarray*}
EH(X',e) & = & \sum_{j=1}^p P(X'=x_j'|e) \cdot H(e,X'=x_j') \enspace .
\end{eqnarray*}
Finally, we choose a question $X^*$ that maximizes the information gain $IG(X',e)$
\begin{eqnarray*}
X^* & = & \operatorname*{arg\,max}_{X' \in \mathcal{X}_s} IG(X',e) \ , \ \mbox{where}\\
IG(X',e) & = & H(e)  - EH(X',e) \enspace .
\end{eqnarray*}

\subsection{INSERTION OF THE SELECTED QUESTION}
The selected question $X^*$ is given to the student and his/her answer is obtained. 
This answer changes the state of variable $X^*$ from unobserved to an observed state $x^*$. 
Next, the question together with its answer is inserted into the vector of evidence $e$. 
We update the probability distributions $P(S_i|e)$ of skill variables with the updated evidence $e$. 
We also recompute the value of entropy $H(e)$.
The question $X^*$ is also removed from $\mathcal{X}_{s}$ 
forming a set of unobserved variables $\mathcal{X}_{s+1}$
for the next step $s$ and selection process can be repeated.

\subsection{ESTIMATING SUBSEQUENT ANSWERS}\label{marker}

In experiments presented in the next section we will use individual models to estimate answers for all subsequent questions 
in $\mathcal{X}_{s+1}$. This is easy since we enter evidence $e$ and perform inference to compute $P(X'=x'|e)$ for all states of 
$X' \in \mathcal{X}_{s+1}$ by invoking the distribute and collect evidence procedures in the BN model.

\section{MODEL EVALUATION}
\label{sec-evaluation}

In this section we report results of tests performed with networks proposed in Section~\ref{sec-models} of this paper. The testing was done by 10-fold cross-validation. For each model we learned the corresponding Bayesian network from $\frac{9}{10}$ of randomly divided data. The model parameters were learned using Hugin's~\citep{hugin} implementation of the EM algorithm. 
The remaining $\frac{1}{10}$ of the dataset served as a testing set. 
This procedure was repeated 10 times to obtain 10 networks for each model type. 

The testing was done as described in Section~\ref{sec-tests}. For every model and for each student from the testing data
we simulated a test run. Collected initial evidence and answers were inserted into the model. 
During testing we estimated answers of the current student based on evidence collected so far. 
At the end of the step $s$ we computed probability distributions $P(X_{i}|e)$ for all unobserved questions 
$X_i \in \mathcal{X}_{s+1}$. Then we selected the most probable state of $X_i$: 
\begin{eqnarray*}
x_i^* & = & \operatorname*{arg\,max}_{x_l}{P(X_{i}=x_l|e)} \enspace .
\end{eqnarray*}
By comparing this value to the real answer $x_i'$ we obtained a success ratio of the response estimation 
for all questions $X_i \in \mathcal{X}_{s+1}$ of test (student) $t$ in step $s$
\begin{eqnarray*}
\operatorname{SR}_s^t & = & \frac{\sum_{X_i \in \mathcal{X}_{s+1}}{I(x_{i}^* = x_{i}')}}{|\mathcal{X}_{s+1}|} \ , \ \mbox{where}\\
I(expr) & = & \left\{ \begin{array}{ll}
1 & \mbox{if $expr$ is true}\\
0 & \mbox{otherwise.}
\end{array}\right.
\end{eqnarray*}
The total success ratio of one model in the step $s$ for all test data (N = 281) is defined as
\begin{eqnarray*}
\operatorname{SR}_s & = & \frac{\sum_{t=1}^N{\mathrm{SR}_s^t}}{N} \enspace .
\end{eqnarray*}
We will refer to the success rate in the step $s$ as to elements of 
$\operatorname{sr} = (\operatorname{SR}_0, \operatorname{SR}_1, \ldots)$, 
where $\operatorname{SR}_0$ is the success rate of the prediction before asking any question.

Table~\ref{tab:BayesianNetworkSuccessRates} shows success rates of proposed networks for selected steps 
$s=0,1,5,15,25,30$. The network ID corresponds to the ID from the Table~\ref{tab:BayesianNetworkModelsOverview}. 
The most important part of the tests are the first few steps, which is because of the nature of CAT.
We prefer shorter tests therefore we are interested in the early progression 
of the model (in this case approximately up to the step $20$). During the final stages of testing we 
estimate results of only a couple of questions which in some cases may cause rapid changes of success rates. Questions which are left to the end of the test do not carry a large amount of information (because of the entropy selection strategy). This may be caused by two possible reasons. The first one is that the state of the question is almost certain and knowing it does not bring any additional information. The second possibility is that the question connection with the rest of the model is weak and because of that it does not change much the entropy of skill variables. In the latter case it is also hard to predict the state of such question because its probability distribution also does not change much with additional evidence.

\begin{table}
	{\centering
		{\small
		\begin{tabular}{lccccccc}
			{ID/Step} & 0 & 1 & 5 & 15 & 25 & 30 \\
			\hline
b2 & 0.714 & 0.761 & 0.766 & 0.778 & 0.798 & 0.835 \\
b2+ & 0.749 & 0.768 & 0.768 & 0.778 & 0.797 & 0.829 \\ 
b3 & 0.714 & 0.745 & 0.776 & 0.803 & 0.843 & 0.857 \\
b3+ & 0.746 & 0.754 & 0.78 & 0.801 & 0.831 & 0.859 \\
b3o & 0.714 & 0.747 & 0.782 & 0.8 & 0.832 & 0.864 \\
b3o+ & 0.747 & 0.761 & 0.785 & 0.799 & 0.83 & 0.865 \\
b2e & 0.715 & 0.73 & 0.767 & 0.776 & 0.781 & 0.768 \\
			
			\hline
			
			n2 & 0.684 & 0.708 & 0.73 & 0.713 & 0.745 & 0.776 \\
n2+ & 0.717 & 0.732 & 0.731 & 0.717 & 0.75 & 0.778 \\
n3 & 0.684 & 0.723 & 0.745 & 0.758 & 0.781 & 0.79 \\
n3+ & 0.684 & 0.724 & 0.743 & 0.757 & 0.77 & 0.776 \\ 
n3o & 0.686 & 0.721 & 0.745 & 0.751 & 0.77 & 0.779 \\
n3o+ & 0.716 & 0.729 & 0.743 & 0.752 & 0.773 & 0.779 \\ 
n2e & 0.684 & 0.699 & 0.735 & 0.738 & 0.737 & 0.715 \\
			\hline
			\end{tabular}}
	\caption{Success ratios of Bayesian network models}
	\label{tab:BayesianNetworkSuccessRates}}
\end{table}

\normalsize
From an analysis of success rates we have identified clusters of models with similar behavior.  
For models with integer valued questions and also for models with Boolean questions 
three clusters of models with similar success ratio emerged:
\begin{itemize}
	\item models with skill variable of $3$ states,
	\item models with skill variable of $2$ states, and
	\item the expert model.
\end{itemize}

\begin{figure*}[htbp]
\includegraphics[width=\textwidth]{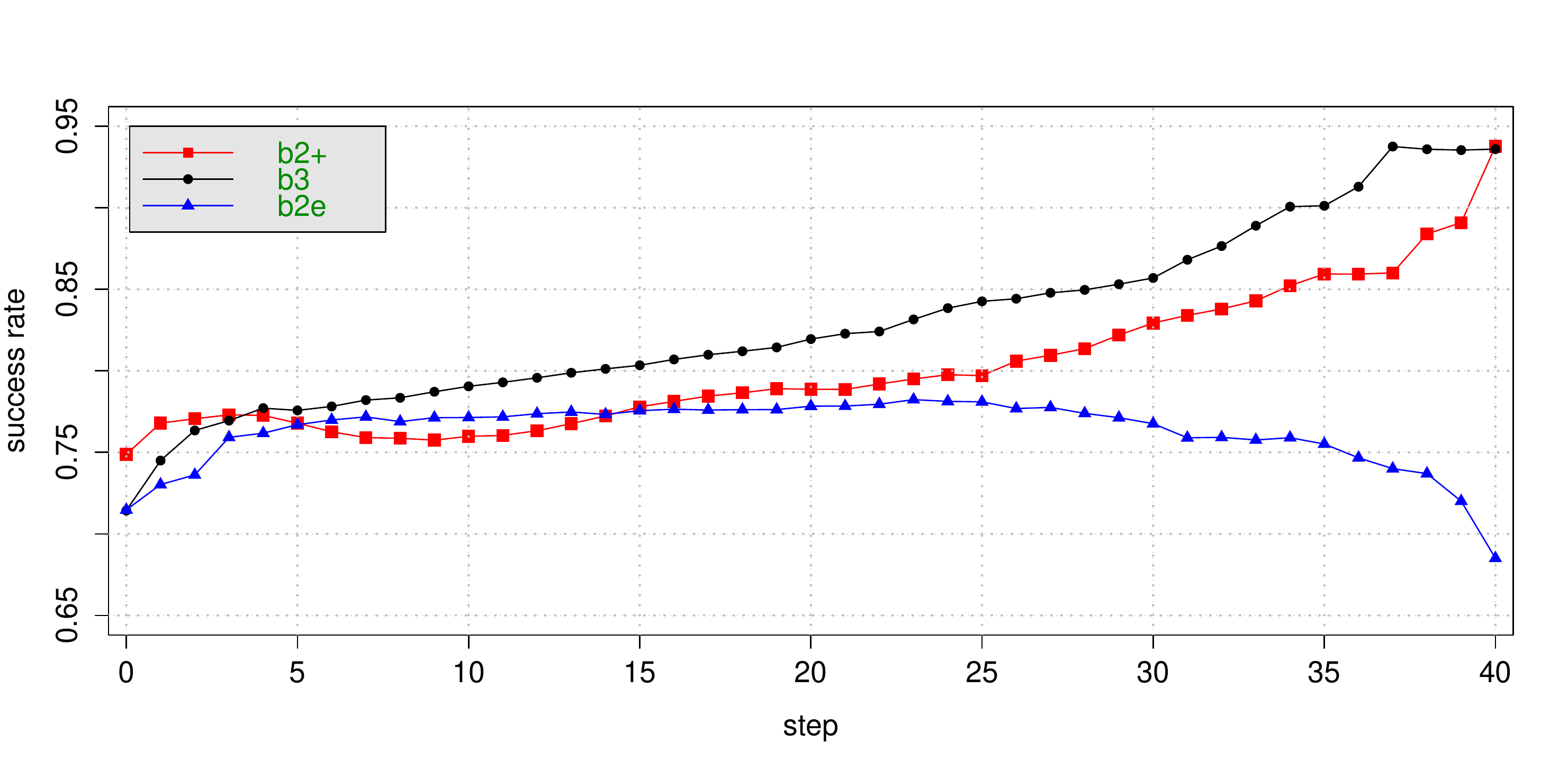}
\caption{Success ratios for models with Boolean questions}
\label{fig:tf}
\end{figure*}

\begin{figure*}[htbp]
\includegraphics[width=\textwidth]{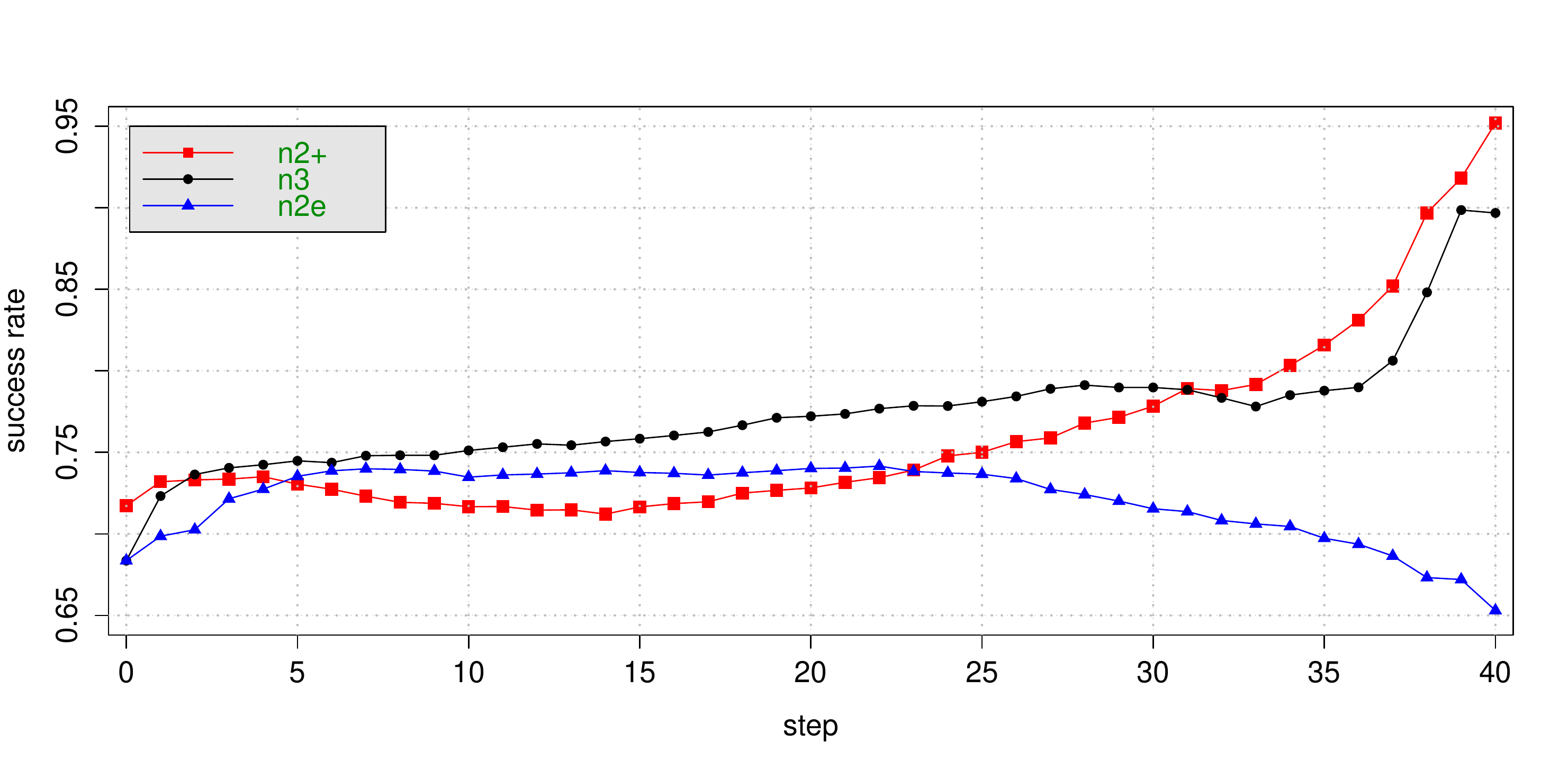}	
\caption{Success ratios for models with integer valued
 questions}
\label{fig:points}
\end{figure*}

We selected the best model from each cluster to display success ratios $\operatorname{SR}_s$ in steps $s$ in Figure~\ref{fig:tf} 
for Boolean questions and in Figure~\ref{fig:points} for integer valued questions. 
We made the following observations:
\begin{itemize}
\item Models with the skill variable with $3$ states were more successful.
\item Models with skill variable with $2$ states were better at the very end of tests, 
but this test stage is is not very important for CAT since the tests usually terminates at early stages as explained above.
\item The expert model achieved medium quality prediction in the middle stage but its prediction ability decreases in the second half of the tests.
\end{itemize}

\begin{figure*}[htbp]
\includegraphics[width=\textwidth]{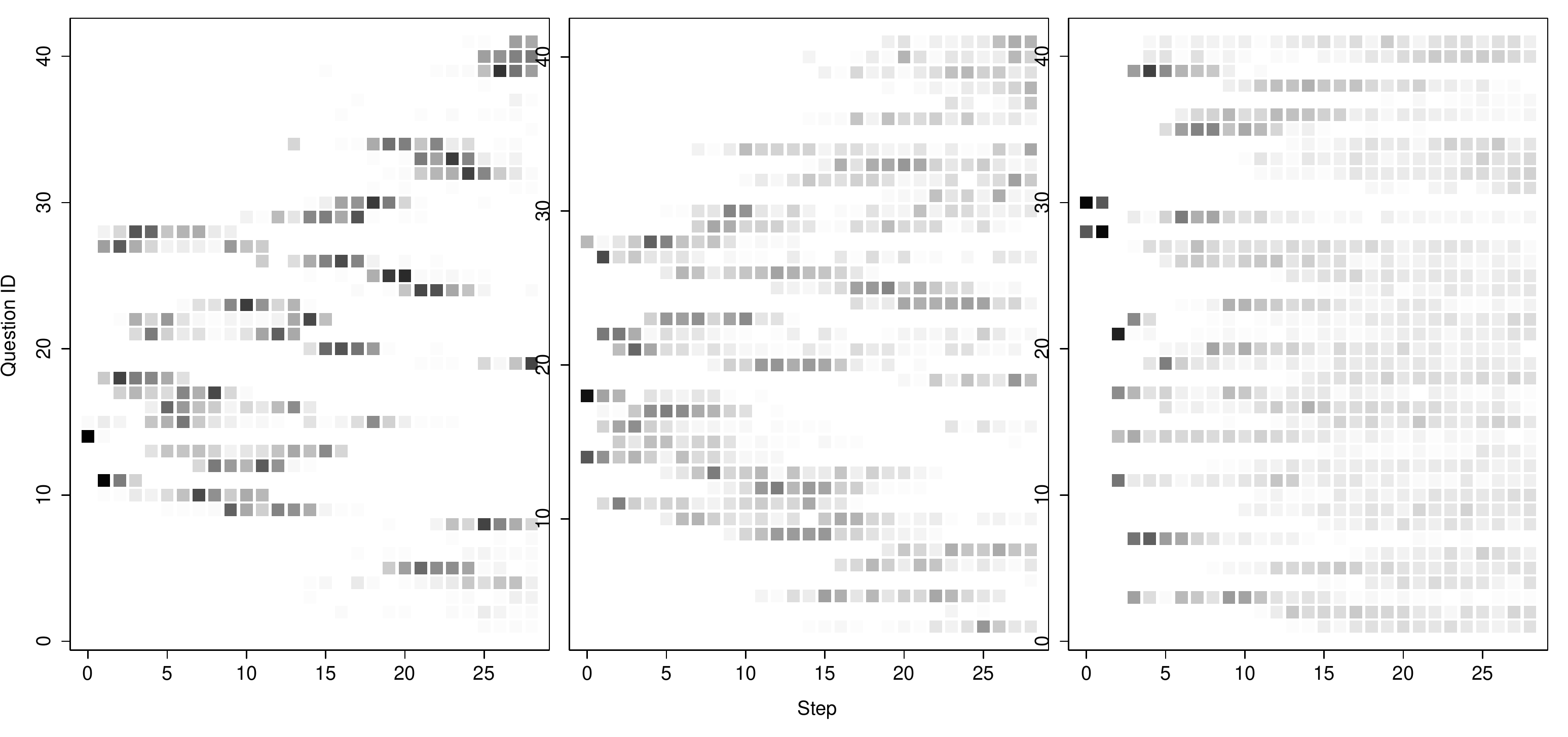}
\caption{\small Relative occurrence of questions (on vertical axis)
 into models with Boolean scale. From left ``b2+'',``b3'',``b2e''}
\label{scatter1}
\end{figure*}

\begin{figure*}[htbp]
\includegraphics[width=\textwidth]{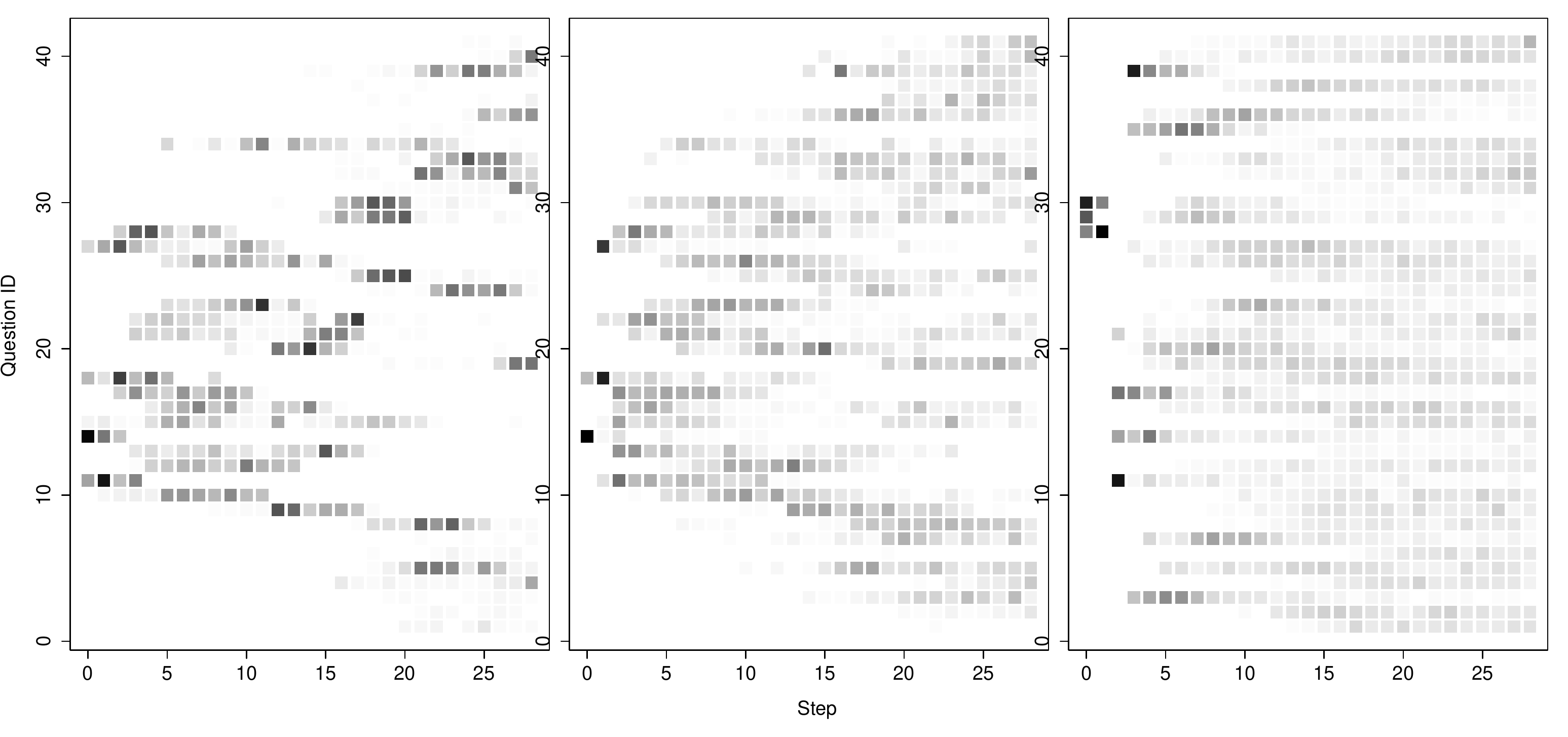}
\caption{\small Relative occurrence of questions (on vertical axis) into models with numeric scale. From left ``n2+'',``n3'',``n2e''}
\label{scatter2}
\end{figure*}

We would like to point out that the distinction between models is basically only by differences of skill variables used in the models. The influence of additional information is visible only at the very beginning of testing. 
As can be seen in the Table~\ref{tab:BayesianNetworkSuccessRates} ``+'' models are scoring better in the initial estimation and then in the first one. After that both models follow almost the same track. In the late stages of the test, models with additional information are estimating worse than their counterparts without information. 
It suggests that models without additional information are able to derive the same information by getting answers 
to few questions (in the order of a couple of steps).

It is easy to observe that the expert model does not provide as good results as other models especially during the second half of the testing. As was stated above the second part of the testing is not as important, nevertheless we have investigated causes for these inaccuracies. The main possible reason for this behavior may be the complexity of this type of model. With seven skill nodes and various connections to question nodes this model contains a significantly higher number of parameters to be fitted. It is possible that our limited learning sample leads to over-fitting. We have explored the conditional probability tables (CPTs) of models used during cross-validation procedure to see how sparse they are. Our observation is shown in the Table~\ref{Tab:sparsity}. The number $\operatorname{AZT}$ is the average of the total number of zeros in cross-validation models for the specific configuration and $\operatorname{AS}$ is the average sparsity of CPTs rows in these models. We can see that in the same type of scales (Boolean or numeric) the sparsity of expert models is significantly higher. This can be improved by increasing data volume or decreasing the model's complexity. This finding is consistent with the above explained possible cause for inaccuracies. In addition we can observe that there is also an increase in sparsity when more skill variables states are introduced. It seems to us as a good idea to further explore the space between one skill variable and seven skill variables as well as the number of their states to provide a better insight into this problem and to draw out more general conclusions.

\begin{table}%
\begin{tabular}{lcccccc}
& b2+ & b3 & b2e & n2+ & n3 & n2e\\
\hline
$\operatorname{AZT}$ & 0.5 & 1.9 & 7.5 & 18.1 & 47.4 &  81.7\\
$\operatorname{AS}$ & 0.002 & 0.006 & 0.026 & 0.047 & 0.081 & 0.121\\
\hline
\end{tabular}
\caption{Avg. number of zeros/sparsity of different models}
\label{Tab:sparsity}
\end{table}

In Figures~\ref{scatter1} and~\ref{scatter2} we compare which questions were often selected 
by the tested models at different stages of the tests.
Figure~\ref{scatter1} is for Boolean questions and Figure~\ref{scatter2} for integer valued questions. 
Only three models (the same as for success ratio plots) were selected because other models share common behavior with others 
from the same cluster. On the horizontal axis there is the step when the question was asked, on the vertical axis are questions by their ID. The darker the cell in the graph the more tests used the corresponding variable in the corresponding time. 
Even though it provides only a rough presentation it is possible to notice different patterns of behavior. Especially, we would like to point out the clouded area of the expert model where it is clear that the individual tests were very different. Expert models are apparently less sure about the selection of the next question. This may be caused by a large set of skill variables which divide the effort of the model into many directions. This behavior is not necessarily unwanted because it provides very different test for every test subject which may be considered positive, but it is necessary to maintain the prediction success rates.

\section{CONCLUSION AND FUTURE RESEARCH}

In this paper we presented several Bayesian network models designed for adaptive testing. 
We evaluated their performance using data from paper tests organized at grammar schools.
In the experiments we observed that:
\begin{itemize}
\item Larger state space of skill variables is beneficial. Clearly, models with 3 states of the hidden skill variable behave better during the most important stages of the tests. Test with hidden variables with more than 3 states are still to be done.
\item Expert model did not score as good as simpler models but it showed a potential for its improvements. The proposed expert model is much more complex than other models in this paper and probably it can improve its performance with more data collected.
\item Additional information provided improves results only during the initial stage. This fact is positive because obtaining such additional information may be hard in practice. Additionally, it can be considered politically incorrect 
to make assumption about student skills using this type of information.
\end{itemize}

In the future we plan to explore models with one or two hidden variables having more than three states,
expert models with skill nodes of more than 2 states, and 
try to add relations between skills into the expert model to improve its performance.
We would also like to compare our current results with standard models used in adaptive testing like the Rash and IRT 
models.

\subsubsection*{Acknowledgements}
The work on this paper has been supported from GACR project n. 13-20012S.

\renewcommand{\refname}{\normalfont\selectfont\normalsize\bfseries References} 
\bibliographystyle{apalike}
\bibliography{mybib}

\end{document}